\documentclass{article}

\PassOptionsToPackage{numbers}{natbib}
\usepackage{lmodern}
\usepackage[preprint]{neurips_2026}
\usepackage[english]{babel}

\usepackage{graphicx} 
\usepackage{hyperref}
\usepackage{amsmath}
\usepackage{amsfonts}
\usepackage{amssymb}
\usepackage{comment}
\usepackage{multirow}
\usepackage{multicol}
\usepackage{ctable}
\usepackage{subcaption}
\usepackage{amsthm}
\theoremstyle{definition}

\usepackage{booktabs}
\usepackage{microtype}
\usepackage{mathtools}
\usepackage[utf8]{inputenc}   
\usepackage[T1]{fontenc}      
\usepackage{xcolor}

\title{Forecasting from Counterfactual Simulator Rollouts: A Sim2Real Evaluation}

\author{%
  Angel Wang \\
  Amazon \\
  \texttt{wangange@amazon.com}
  \And
  Dominique Perrault-Joncas \\
  Amazon \\
  \texttt{joncas@amazon.com}
  \And
  Alvaro Maggiar \\
  Amazon \\
  \texttt{maggiara@amazon.com}
  \And
  Dean Foster \\
  Amazon \\
  \texttt{foster@amazon.com} \\
  \And
  Carson Eisenach \\
  Amazon \\
  \texttt{ceisen@amazon.com}
}

\begin{document}

\maketitle

\begin{abstract}
Deploying a new decision policy creates a cold-start problem for prediction models whose targets depend on the policy’s actions: historical observations reflect earlier policies, while real observations under the new policy are not yet available. Simulation offers a way to address this gap by rolling out the target policy across counterfactual scenarios and using the resulting trajectories to learn how the system responds to those controls. The simulation-to-reality (Sim2Real) transfer of this simulator-trained model can then be backtested by evaluating it against real observations from past deployments. Using two real-world inventory-control deployments, we evaluate this process from three angles: simulator fidelity, zero-shot transfer to real behavior, and adaptation as real target-policy observations accumulate. The simulator-trained forecaster achieves lower point-estimate mean absolute percentage error (MAPE) than the same architecture trained on historical real data, reducing MAPE by 1.2–3.1 percentage points in Study 1 and 12.5–18.7 points in Study 2. After deployment, lightweight calibration using early real observations further reduces error by up to 2.5 percentage points. These results provide empirical evidence that simulator-generated counterfactual data can support cold-start forecasting under a new policy, and the resulting model can be further refined as real deployment data become available.
\end{abstract}

\section{Introduction}

Decision policies in real-world systems are increasingly developed offline and evaluated in simulation before deployment~\citep{madeka2022deep,eisenach2024ncc,dulacarnold2021challenges}. This raises the classical simulation-to-reality (Sim2Real) question of whether behavior learned or evaluated in simulation transfers to the real system~\citep{tobin2017domain,kadian2020sim2real}.

A related problem arises when deployment also relies on predictive models whose targets depend on the policy's actions -- queue lengths under scheduling, traffic conditions under routing, power demand under load control, or future inventory under replenishment. Historical observations reflect previously operated policies, while deploying a new one can induce a different distribution of states, actions, and resulting system responses. Prediction models whose targets depend on these responses therefore face a cold-start problem: real observations under the target policy do not yet exist. Simulation can address this gap by rolling out the target policy under counterfactual scenarios and generating on-policy trajectories before corresponding real observations are available.

This shifts the Sim2Real question from whether a simulator-trained policy transfers to whether a prediction model learned from simulated target-policy trajectories transfers to real behavior. We evaluate this process from three angles: simulator fidelity, zero-shot predictive transfer, and adaptation as real target-policy observations become available.

We study this setting in large-scale inventory control using an exogenous interactive decision process (Exo-IDP) simulator~\citep{madeka2022deep,eisenach2024ncc} and two historical deployments of different inventory-ordering policies at a major retailer. Product-level replenishment decisions jointly determine aggregate inbound volume, which must be coordinated with shared inbound processing capacity. Capacity planning therefore requires forecasting how aggregate inbound will respond to system-level coordination signals. We use the simulator to roll out the deployed policy over a range of counterfactual coordination signals and train an aggregate forecaster of the resulting inbound utilization~\citep{wang2026population}. We then backtest the simulator-trained forecaster against realized deployment outcomes and evaluate whether it can be further refined as real target-policy observations become available. Across both deployments, the simulator-trained forecaster achieves lower point-estimate error than the same architecture trained on historical real data, while online calibration further reduces error.
\section{Related Work} \label{sec:related}

\paragraph{Sim2Real transfer in reinforcement learning.}
Sim2Real transfer has been widely studied in robotics and reinforcement learning (RL), where the central challenge is bridging the ``reality gap'' between simulated and physical environments~\citep{jakobi1995evolutionary,tobin2017domain,peng2018sim,tan2018sim}. Prior work studies, among other settings, training policies under randomized simulation~\citep{tobin2017domain,peng2018sim}, adapting simulation-trained controllers or dynamics models with real data~\citep{tan2018sim,nagabandi2018neural}, and measuring whether simulated performance predicts real performance~\citep{kadian2020sim2real,muratore2021assessing}. Our setting differs in the object being transferred: the decision policy is fixed, and we ask whether a prediction model learned from simulator-generated trajectories under that policy transfers to realized system behavior.

\paragraph{Synthetic training data and backtest evaluation.}
Synthetic data have been used before to train forecasting models for broad generalization, including pretraining on simulated time series~\citep{dooley2023forecastpfn,ansari2024chronos}. Our approach extends this idea by training the forecaster on policy-conditioned simulated data, which is then validated against realized outcomes from the corresponding deployment. Our evaluation is retrospective~\citep{mizrahi2021embracing,cerqueira2020evaluating}: the forecaster is trained using information available before the corresponding deployment outcomes and evaluated against subsequently realized outcomes. Because those outcomes are observed under the deployed policy, this evaluation does not require the counterfactual corrections used in off-policy evaluation~\citep{levine2020offline,jiang2016doubly}.

\section{Backtesting Sim2Real Transfer of Simulator-Trained Forecasts}
\label{sec:setup}

\subsection{Exo-IDP Simulator and Counterfactual Rollouts}
\label{subsec:task}

\paragraph{Exo-IDP simulator.}
Let $\mathcal M$ denote the real decision process and $\hat{\mathcal M}$ an approximate
simulator of that process. Our setting follows the Exogenous Interactive Decision Process
(Exo-IDP) formulation~\citep{madeka2022deep,eisenach2024ncc}. For each agent $i$,
the local state separates into an exogenous process $\xi_t^i$, which evolves independently of the
agent's actions, and an endogenous state $y_t^i$, which depends on past actions:
\[
x_t^i=(y_t^i,\xi_t^i),
\qquad
a_t^i \sim \pi(\cdot\mid x_t^i),
\qquad
y_{t+1}^i
\sim
P(\cdot\mid y_t^i,a_t^i,\xi_t^i).
\]
An Exo-IDP simulator replays historical realizations of the exogenous process while evolving
the endogenous state using approximate dynamics $\hat P$. The same exogenous trajectory can
therefore be used to simulate the system response under counterfactual actions.

\paragraph{Inventory instantiation.}
In our inventory-control application, each agent is a product managed by a replenishment policy,
and $a_t^i$ is its order quantity. Demand $D_t^i$, prices, and
vendor lead times $v_t^i$ are exogenous, while inventory, open orders, and inbound receipts are
endogenous. The endogenous dynamics are governed by the following inventory-flow relationships:
orders placed in previous periods arrive according to their realized lead times,
\begin{equation}
J_t^i
=
\sum_{0\leq k<t}
a_k^i\,\mathbf{1}_{\{v_k^i=t-k\}},
\label{eq:inbound_dynamics}
\end{equation}
where $J_t^i$ denotes inbound receipts for product $i$. Inventory evolves as
\begin{equation}
I_{t^-}^i = I_{t-1}^i + J_t^i,
\qquad
s_t^i = \min(D_t^i,I_{t^-}^i),
\qquad
I_t^i = (I_{t^-}^i-D_t^i)^+,
\label{eq:inventory_dynamics}
\end{equation}
where $s_t^i$ denotes realized sales.

Equations~\eqref{eq:inbound_dynamics} and~\eqref{eq:inventory_dynamics} produce the counterfactual
rollouts used below to study how the fixed replenishment policy responds to alternative
capacity-control inputs. Full simulator construction details are given in
Appendix~\ref{app:exo_idp}.

\subsection{Learning an Aggregate Response Forecaster}
\label{subsec:forecasting_task}

\paragraph{Shared capacity and coordination signal.}
In the capacity-constrained inventory-control problem, the shared resource is inbound capacity:
aggregate inbound utilization
\[
J_t=\sum_i J_t^i
\]
must satisfy $\mathbb E[J_t]\leq G_t$ for the available capacity $G_t$. This aggregate utilization
is the quantity we forecast.
Following the Lagrangian coordination formulation of~\citet{eisenach2024ncc}, the dual variable
$\lambda_t \ge 0$ acts as a broadcast per-unit cost on inbound utilization: higher costs discourage ordering
and thereby reduce future inbound. The replenishment policy
therefore responds to both its local state and the shared cost signal,
\[
a_t^i\sim\pi(\cdot\mid x_t^i,\lambda_t).
\]
Through the inventory dynamics above, these product-level policy responses jointly determine
aggregate inbound $J_t$. Capacity planning requires evaluating how candidate cost signals change
this aggregate utilization relative to the desired capacity plan. We therefore learn a response
forecaster that predicts aggregate inbound under a proposed cost trajectory.

\paragraph{Response forecaster.}
Let $X_t$ denote the forecast-origin representation constructed from local-state and action
trajectories over an $H$-week lookback window,
\[
X_t
=
\Phi\!\left(
\{x_\tau^i,a_\tau^i\}_{i,\tau=t-H}^{t-1}
\right).
\]
Given $X_t$ and a proposed cost trajectory $\lambda_{t:t+L}$, the response forecaster predicts
aggregate inbound over horizon $L$,
\begin{equation}
\hat J_{t:t+L}
=
\psi_\theta(X_t,\lambda_{t:t+L}).
\label{eq:response_task}
\end{equation}
This forward map from proposed control inputs to aggregate utilization corresponds to the primal
coordination interface of~\citet{wang2026population}.

\paragraph{Training on rollouts.}
We train $\psi_\theta$ by supervised learning on policy-conditioned simulated rollouts. We generate
capacity-control scenarios represented by cost trajectories $\lambda_{0:T}$ and apply each trajectory
to the target policy in $\hat{\mathcal M}$. The simulator evolves the resulting actions and
endogenous states and records aggregate inbound, yielding supervised examples
\[
(X_t,\lambda_{t:t+L},J_{t:t+L}).
\]
We use the aggregate multi-horizon forecasting architecture described in
Appendix~\ref{app:architecture}, which also provides feature and rollout-generation details.

\subsection{Backtesting on Historical Deployments}
\label{subsec:protocol}
\paragraph{Backtest protocol.}
The central question is whether a response learned from an imperfect simulator transfers to the
real system. Completed deployments make this directly testable: realized target-policy outcomes
provide ground truth for a forecaster trained on counterfactual data.
For a deployed policy $\pi$, let $D^\pi_{\hat{\mathcal M}}$ denote the distribution of forecasting
examples $(X,\lambda,J)$ generated by simulator rollouts under that policy, and let
$D^\pi_{\mathcal M}$ denote the distribution realized under $\pi$ in the corresponding real
deployment. The simulator-trained forecaster is
$\hat\psi_{\mathrm{sim}} = \mathrm{Learn}(D^\pi_{\hat{\mathcal M}})$,
and its zero-shot Sim2Real risk is the expected loss
$\mathbb E_{(X,\lambda,J)\sim D^\pi_{\mathcal M}}[\ell(\hat\psi_{\mathrm{sim}}(X,\lambda),J)]$.

\paragraph{Cold-start benchmark.}
We train the same forecasting architecture on historical real
observations collected under earlier policies $\pi_0$,
$\hat\psi_{\mathrm{hist}} = \mathrm{Learn}(D^{\pi_0}_{\mathcal M})$,
and evaluate it on target-policy deployment outcomes.

\paragraph{Online calibration.}
After deployment, real target-policy observations become available sequentially. We evaluate
whether these observations can refine the simulator-trained forecast using an online calibration
(an expanding-window linear regression). At forecast origin $t$, let
$\hat J_\tau^{\mathrm{sim}} = \hat\psi_{\mathrm{sim}}(X_\tau,\lambda_{\tau:\tau+L})$
denote an earlier simulator-trained forecast. A single pair $(\alpha_t,\beta_t)$, shared across
horizons, is estimated from all horizon-level forecast--outcome pairs observed before $t$,
\begin{equation}
(\hat\alpha_t,\hat\beta_t)
=
\arg\min_{\alpha,\beta}
\sum_{\tau<t}\,
\sum_{h\in\mathcal H_{\tau,t}}
\left(
J_{\tau+h}-\alpha-\beta\hat J_{\tau,h}^{\mathrm{sim}}
\right)^2,
\label{eq:calibration_fit}
\end{equation}
where $\mathcal H_{\tau,t}$ indexes the horizons of origin $\tau$ whose outcomes are observed by
$t$; calibration begins after the first four deployment weeks.
We then apply the fitted calibration to the subsequent forecast,
$\hat J_t^{\mathrm{cal}} = \hat\alpha_t + \hat\beta_t \hat J_t^{\mathrm{sim}}$.

\section{Results}
\label{sec:validation}

We evaluate on two completed \emph{production} deployments of inventory-ordering policies, run in
different years under different policies: \textbf{Study~1} is a year-long randomized deployment of an
RL ordering policy covering approximately 20K products, and \textbf{Study~2} is a later three-month
deployment of a different policy covering approximately 100K products. Both are real deployments
whose realized outcomes are observed, enabling retrospective evaluation. Forecast accuracy is
evaluated at the aggregate level using mean absolute percentage error (MAPE) at one-, two-, and
three-week horizons, with weeks 4--10 grouped as a longer planning horizon; reported values average the absolute
percentage error of the aggregate forecast across weekly forecast origins.

\subsection{Fidelity of Simulator-Generated Training Data}
\label{subsec:fidelity}

The simulator reproduces the forecasting target to within roughly $12$--$13\%$: rolling out the
deployed policy one week at a time from realized states, while replaying each evaluation window's
realized exogenous sequences, yields aggregate-inbound MAPE of $13.2 \pm 2.4\%$ in Study~1 and
$11.8 \pm 2.6\%$ in Study~2 against realized outcomes. Because this diagnostic conditions on
realized information unavailable at forecast time, it isolates the simulator's endogenous dynamics
and serves as a reference point rather than a deployable forecast.

We additionally examine fidelity at the product--week level. For each simulated action or state
variable $Y$, we regress the corresponding realized value on its simulated counterpart without intercept,
\[
Y_t^i = \beta_Y \hat Y_t^i + \varepsilon_t^i.
\]
A slope $\beta_Y = 1$ indicates calibration in scale, while $R^2$ measures cross-sectional
agreement. Order quantity and inbound exhibit larger calibration gaps, whereas inventory and sales
drain are more closely calibrated (Table~\ref{tab:fidelity_components}). Calibration is also
stronger for products that actually operated under the deployed policy than for those under the
baseline, consistent with measured fidelity being policy-conditional (Appendix~\ref{app:fidelity_diagnostics}).

\begin{table}[h]
\centering
\caption{\small Component-level trajectory calibration (Study~1): realized values regressed on
simulated values at the product--week level, with slope $1$ indicating perfect calibration. Order and
inbound exhibit larger calibration gaps; inventory and sales drain are better calibrated.}
\label{tab:fidelity_components}
\footnotesize
\begin{tabular}{lcc}
\toprule
\textbf{Component} & \textbf{Calibration slope (95\% CI)} & $R^2$ \\
\midrule
Order quantity & $[0.70,\,0.71]$ & $0.24$ \\
Inbound        & $[0.67,\,0.67]$ & $0.47$ \\
Inventory      & $[0.90,\,0.91]$ & $0.81$ \\
Sales drain    & $[0.99,\,1.00]$ & $0.92$ \\
\bottomrule
\end{tabular}
\end{table}

\subsection{Zero-Shot Transfer of Simulator-Trained Forecasts} \label{subsec:transfer}

Before testing transfer to reality, we verify that the simulated cost-response relationship is
learnable: over the same evaluation windows, the architecture achieves $9.1$--$11.9\%$ MAPE on
held-out \emph{simulated} inbound, comparable to models trained and evaluated on real data
(Appendix~\ref{app:sim2real_detail}), confirming that the simulated trajectories provide a usable
supervised training signal.

We then evaluate whether this learned response transfers to realized deployment outcomes. For each
study, the forecaster is trained only on simulator rollouts under the target policy
(Section~\ref{subsec:forecasting_task}). The historical-data baseline uses
the same forecasting architecture trained on real observations collected under earlier policies.

Simulator-trained forecasters transfer: they achieve lower point-estimate MAPE than the off-policy
historical baseline at every reported horizon in both studies (Table~\ref{tab:transfer}), with
reductions of $1.2$--$3.1$ percentage points in Study~1 and $12.5$--$18.7$ points in Study~2
($8$--$21\%$ and $45$--$67\%$ in relative terms). Notably, the forecaster attains
accuracy comparable to the direct-fidelity diagnostic
of Section~\ref{subsec:fidelity} while using only information available at the forecast origin.
This is consistent with counterfactual rollouts providing training data that reflect the target
policy's response dynamics across the control range, whereas historical observations are limited to
previously realized policies and controls; the larger gains in Study~2 are consistent with a larger
shift between the prior operating regime and the deployed policy.
\begin{table}[t]
\centering
\caption{\small Zero-shot transfer: MAPE (mean $\pm$ 95\% CI) on realized deployment outcomes for the same forecaster architecture trained on simulator rollouts under the deployed policy versus historical real data from prior policies (off-policy), by forecast horizon. CIs are 95\% normal-approximation intervals across weekly forecast origins, with one error observation per origin and horizon.}
\label{tab:transfer}
\footnotesize
\setlength{\tabcolsep}{4pt}
\begin{tabular}{llrrrr}
\toprule
\textbf{Study} & \textbf{Training source} & \textbf{1 week} & \textbf{2 weeks} & \textbf{3 weeks} & \textbf{4--10 weeks} \\
\midrule
Study~1 & Simulator rollouts & $13.3\pm 2.7$ & $11.7\pm 2.8$ & $13.2\pm 3.1$ & $14.8\pm 2.8$ \\
Study~1 & Historical (off-policy) & $15.0\pm 2.9$ & $14.8\pm 3.1$ & $15.2\pm 3.1$ & $16.0\pm 3.1$ \\
\midrule
Study~2 & Simulator rollouts & $12.7\pm 3.6$ & $15.3\pm 7.4$ & $9.4\pm 3.8$ & $11.5\pm 3.4$ \\
Study~2 & Historical (off-policy) & $25.6 \pm 5.2$ & $27.8 \pm 6.3$ & $28.1 \pm 6.4$ & $25.2 \pm 5.7$ \\
\bottomrule
\end{tabular}
\end{table}

\subsection{Adaptation with Real Observations}
\label{subsec:adaptation}

As real on-policy observations accumulate, we can also update the simulator-trained forecaster using
the online calibration of Section~\ref{subsec:protocol}: the regression is refit at each forecast
origin using all outcomes available up to that time and applied to the subsequent forecast.

Online calibration further reduces error: point-estimate MAPE improves at multiple horizons, with
the largest reduction of $2.5$ percentage points ($9.4\%$ to $6.9\%$) in Study~2
(Table~\ref{tab:adaptation});
gains are concentrated at horizons of three weeks or less and are not uniform across settings. Calibration thus complements rather than
replaces the simulator-based cold start: rollouts provide the pre-deployment forecaster, and early
real observations refine it.

\begin{table}[t]
\centering
\caption{\small Calibration with early real observations: MAPE (mean $\pm$ 95\% CI) for the simulator-trained forecaster, zero-shot versus after regression calibration on early deployment observations, by forecast horizon. Zero-shot and calibrated variants are evaluated on the same forecast origins as Table~\ref{tab:transfer}, whose interval definition applies. Calibrated historical-data baselines are reported in Appendix~\ref{app:sim2real_detail}.}
\label{tab:adaptation}
\footnotesize
\setlength{\tabcolsep}{4pt}
\begin{tabular}{llrrrr}
\toprule
\textbf{Study} & \textbf{Variant} & \textbf{1 week} & \textbf{2 weeks} & \textbf{3 weeks} & \textbf{4--10 weeks} \\
\midrule
Study~1 & Zero-shot & $13.3\pm 2.7$ & $11.7\pm 2.8$ & $13.2\pm 3.1$ & $14.8\pm 2.8$ \\
Study~1 & +Calibration & $11.8\pm 2.6$ & $11.2\pm 3.1$ & $13.1\pm 3.6$ & $14.8\pm 3.2$ \\
\midrule
Study~2 & Zero-shot & $12.7\pm 3.6$ & $15.3\pm 7.4$ & $9.4\pm 3.8$ & $11.5\pm 3.4$ \\
Study~2 & +Calibration & $12.6 \pm 3.6$ & $13.0 \pm 7.1$ & $6.9 \pm 3.1$ & $9.7 \pm 2.9$ \\
\bottomrule
\end{tabular}
\end{table}

\section{Discussion and Limitations} \label{sec:discussion}

We investigate whether simulator-generated counterfactual data can address the forecasting cold start created by deployment of a new decision policy. Across two large-scale supply-chain deployments, the simulator exhibits measurable discrepancies from realized behavior, yet forecasters trained on target-policy simulator rollouts transfer to real outcomes and achieve lower point-estimate error than the same architecture trained on historical prior-policy data. As real target-policy observations become available, lightweight calibration further reduces error, especially at shorter horizons. Useful downstream transfer thus does not require perfect direct simulator fidelity; fidelity metrics alone are an incomplete measure of a simulator's value as a training-data generator, and characterizing which fidelity properties govern transfer remains open.

More broadly, our results support simulation as a source of target-policy training data before real observations exist: the simulator can expose the target policy to a broad range of control scenarios and generate the corresponding system responses -- coverage that historical data, limited to the combinations that actually occurred, cannot provide. We validated the approach on two deployments of one supply-chain simulator; the evidence so far is application-specific, and extending the evaluation to additional simulators and domains is a natural next step.

\newpage

\bibliographystyle{ims_nourl_eprint}
\bibliography{external}

\newpage
\appendix

\section{Exo-IDP Simulator and Historical Replay} \label{app:exo_idp}
\subsection{Interactive Decision Process Formulation}
\label{app:idp_formulation}

This appendix instantiates the Exo-IDP setting of
Section~\ref{sec:setup} in the supply-chain simulator used to generate
counterfactual rollout data. We consider discrete time steps $t=0,\ldots,T$,
where $S_t$ is the active product population and each agent $i\in S_t$
corresponds to a product governed by a local replenishment policy. The shared
resource in this instantiation is inbound processing capacity, so the aggregate
inbound utilization $J_t$ of the main text is realized as aggregate
\emph{inbound} volume -- the terms are used interchangeably in the appendices --
and $\lambda_t$ is the per-period inbound cost signal.

\paragraph{Exogenous processes.}
For each product $i$ and week $t$, the exogenous state series is
\[
\xi_t^i = (D_t^i,\;p_t^i,\;c_t^i,\;v_t^i),
\]
where the exogenous variables are demand, sales price, ordering cost, and
vendor lead time. Importantly, these variables are outside the control of the
ordering actions; thus exogenous to the decision process. To be precise,
$D_t^i$ denotes the \emph{unconstrained} demand -- demand not affected by
historical stock availability -- and $v_t^i$ is the lead time incurred when an
order is placed. Because $\xi_{0:T}^i$ evolves independently of the policy, its
historical realizations can be replayed under counterfactual actions; this
property enables counterfactual rollout generation under alternative actions
and coordination signals.

\paragraph{Endogenous states.}
The endogenous state $y_t^i$ collects the inventory variables -- on-hand
inventory, open orders, and arrivals -- which depend on past ordering actions.
Unlike the exogenous processes, these cannot be replayed: under a
counterfactual policy they must be simulated forward from the dynamics below.
The full product state is $x_t^i = (y_t^i,\;\xi_t^i)$.

\paragraph{Control process.}
At each time $t$, each product observes its local state and the broadcast cost
signal $\lambda_t$ and selects an order quantity according to a local
replenishment policy:
\[
a_t^i \sim \pi(\cdot \mid x_t^i, \lambda_t).
\]

\paragraph{Transition dynamics.}
The endogenous state evolves as
\[
y_{t+1}^i \sim P(\cdot \mid y_t^i,\;a_t^i,\;\xi_t^i).
\]
Orders placed in previous periods arrive during period $t$ according to their
realized lead times:
\[
J_t^i = \sum_{0\le k<t} a_k^i\,\mathbf{1}_{\{v_k^i = t-k\}},
\]
where $v_k^i$ is the realized vendor lead time for the order placed at time $k$.
Beginning-of-period inventory after receipts is
\[
I_{t-}^{i} = I_{t-1}^i + J_t^i.
\]
End-of-period inventory after demand fulfillment is
\[
I_t^i = (I_{t-}^{i} - D_t^i)^+,
\]
and realized sales are
\[
s_t^i = \min(D_t^i,\;I_{t-}^{i}).
\]
Aggregate inbound utilization across the active population is
\[
J_t = \sum_{i\in S_t} J_t^i.
\]

\paragraph{Reward.}
The per-agent utility $U_t^i(x_t^i,a_t^i)$
is instantiated as a revenue-minus-ordering-cost reward:
\[
R_t^i = p_t^i s_t^i - c_t^i a_t^i.
\]
The unconstrained inventory-control objective is
\[
\max_{\pi}\;\mathbb{E}\!\left[\sum_{t=0}^T \sum_{i\in S_t} R_t^i\right].
\]

\paragraph{Cost signal.}
\label{app:constrained_opt}
The system broadcasts a nonnegative per-unit inbound cost $\lambda_t$, and the
fixed local policy responds as $a_t^i\sim\pi(\cdot\mid x_t^i,\lambda_t)$:
higher costs reduce ordering, lower costs relax it. The response forecaster
therefore treats the proposed cost trajectory $\lambda_{t:t+L}$ as an input.
The deployment-time optimization of $\lambda_t$ is outside the scope of this paper;
Appendix~\ref{app:capacity_curve_sampling} describes how capacity scenarios are mapped to cost
trajectories for training-data generation.

\subsection{Simulator Construction}
\label{app:gym_construction}

Given historical data $\mathcal{D}$ collected under a historical operating policy
$\pi_{\mathcal{D}}$, we construct the simulator of the main text as
\[
\hat{\mathcal{M}} = (\boldsymbol{\xi}_{\mathcal{D}},\;\hat{P},\;\hat{R}),
\]
where $\boldsymbol{\xi}_{\mathcal{D}}$ are replayed historical exogenous
sequences, and $\hat{P}$, $\hat{R}$ are learned approximations of the
endogenous transition dynamics $P$ and per-agent reward $R_t^i$ defined in
Section~\ref{app:idp_formulation}. Conditional on $\boldsymbol{\xi}_{\mathcal{D}}$, endogenous
states are simulated through $\hat{P}$, enabling counterfactual rollouts under
alternative capacity plans and coordination signals.

Following \citet{madeka2022deep}, Exo-IDP simulators constructed from historical
data admit counterfactual policy-evaluation guarantees under conditions including replayable exogenous trajectories and sufficiently calibrated endogenous dynamics. For the setting here, this requires the simulator to reproduce the response induced by the deployed policy with enough fidelity to
provide a useful training signal; Sections~\ref{subsec:fidelity} and~\ref{app:fidelity_diagnostics} evaluate this fidelity empirically under the deployment conditions.

\section{Response Forecaster and Training-Data Generation} \label{app:architecture}

\subsection{Base Architectures}

The main text writes the response forecaster compactly as
$\hat{J}_{t:t+L} = \psi_\theta(X_t,\lambda_{t:t+L})$ (Eq.~\ref{eq:response_task}), with $X_t$
constructed from local-state and action trajectories over the lookback window. In this appendix, we
expand $X_t$ into per-product histories over the active product set $S_t$ (defined in
Appendix~\ref{app:exo_idp}) and aggregate/global histories; note that the implemented $X_t$
additionally includes known future covariates over the forecast horizon (e.g., holiday distances and
forecasted demand, listed below). We then specify the encoder and decoder for the two forecaster
classes evaluated in Appendix~\ref{app:sim2real_detail}.

\paragraph{Input.}
We use $H{=}64$ historical steps and $L{=}26$ forecast horizon.
\begin{itemize}
  \item $x^{i}_{t-H:t} \in \mathbb{R}^{H \times d^i}$: per-product historical features for product $i \in S_t$
  \item $x^{(g)}_{t-H:t} \in \mathbb{R}^{H \times d_g}$: global historical features (aggregated across products)
  \item $x^{(f)}_{t:t+L} \in \mathbb{R}^{L \times d_f}$: future known features (e.g., holidays, forecasted demand)
  \item Capacity signal: $\lambda_{t:t+L} \in \mathbb{R}^{L}_{\ge 0}$, the proposed cost trajectory
\end{itemize}

\paragraph{Encoder.}
Each historical sequence is mapped to a fixed-size embedding by a stacked dilated
causal CNN:
\[
e_t^i = \mathrm{Enc}_\theta(x^i_{t-H:t}), \quad \text{dilation rates } [1,2,4,8,16,32].
\]
Applied to the global history $x^{(g)}_{t-H:t}$ this yields $e^{(g)}_t$; applied to
each product history $x^i_{t-H:t}$ this yields $e_t^i \in \mathbb{R}^{d_e}$, where
$d_e{=}32$ is the encoder output dimension.

\paragraph{Decoder.}
For response prediction, the decoder maps the relevant state representation and proposed
cost trajectory to a multi-horizon utilization forecast. Global Aggregate uses aggregate features, while Bottom-Up sums independent per-agent predictions over $S_t$:
\[
\hat{J}_{t:t+L} =
\begin{cases}
  \mathrm{Dec}_\theta(e^{(g)}_t,\; x^{(f)}_{t:t+L},\; \lambda_{t:t+L}),
    & \text{Global Aggregate}, \\[4pt]
  \displaystyle\sum_{i\in S_t}\mathrm{Dec}_a(e_t^i,\; x^{(f)}_{t:t+L},\; \lambda_{t:t+L}),
    & \text{Agent Bottom-Up}.
\end{cases}
\]
Concretely, each decoder processes the cost trajectory and future features
through another dilated CNN, concatenates the result with the remaining inputs, and applies
a two-layer ELU MLP to produce its horizon-level output.

\subsection{Input Features}

The forecasters consume the following features.
Historical features span $[t{-}H, t)$; future features span $[t, t{+}L)$.

\begin{itemize}
  \item Historical state (aggregate, $[t{-}H,\,t)$):
    order quantities $\sum_i a^i$,
    inventory $\sum_i I^i$,
    availability-corrected demand $\sum_i \hat{D}^i$,
    inbound $\sum_i J^i$.
  \item Historical state (per-product $i \in S_t$, $[t{-}H,\,t)$):
    order quantity $a^i$, inventory $I^i$, demand $\hat{D}^i$, inbound $J^i$,
    unit economics (price, cost).
  \item Forecasted aggregate ($[t,\,t{+}L)$, $L$ weeks):
    mean demand $\hat{D}_{t:t+L} = \sum_i \hat{D}^i_{t:t+L}$;
    projected inventory after expected drain
    $\sum_i\!\bigl(I^i_t - \textstyle\sum_{\tau=t}^{t+L-1} \hat{D}^i_\tau\bigr)$.
  \item Time-series (global): distance to public holidays.
  \item Capacity cost (global): proposed cost trajectory $\lambda_{t:t+L}$.
\end{itemize}

\subsection{Training-Data Generation} \label{app:capacity_curve_sampling}

\begin{figure}[htbp]
\centering
\includegraphics[width=0.8\textwidth]{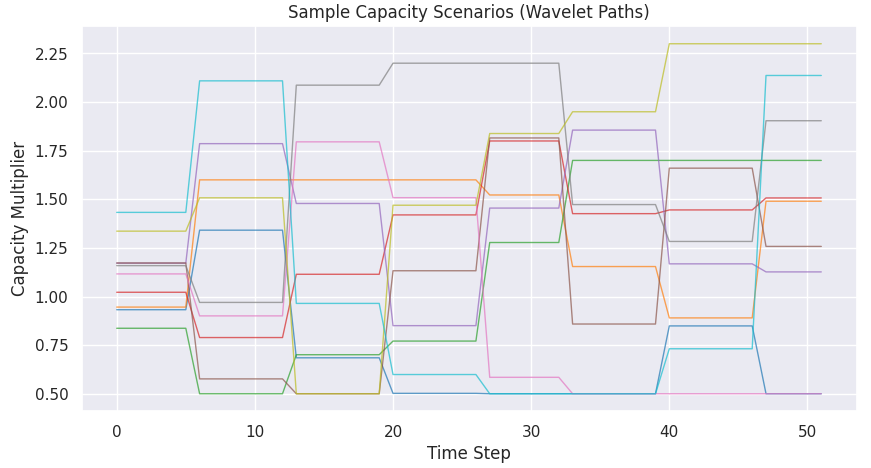}
\caption{\small Example capacity target trajectories generated by the wavelet sampler using a truncated Haar wavelet basis.}
\label{fig:wavelet_sampler}
\end{figure}

The response forecaster is trained on cost trajectories and the aggregate
utilization they induce in closed-loop simulator rollouts, so training requires a
distribution over cost trajectories $\lambda_{0:T}$ that covers the
capacity-control problems the deployment will face. We use a constraint-space
sampling strategy. First, target capacity trajectories $G^{(n)}_{0:T}$ are
generated using a truncated Haar wavelet basis, following~\citep{eisenach2024ncc}:
random Gaussian coefficients produce a diverse family of piecewise-constant
capacity paths with controlled temporal variation, capturing realistic on/off
capacity shifts (Fig.~\ref{fig:wavelet_sampler}). Each sampled path is rescaled to
the aggregate utilization scale of the target population; in our supply-chain
setting, this corresponds to aggregate inbound volume. Second, each sampled plan
is mapped to a broadcast cost trajectory $\lambda^{(n)}_{0:T}$ by a trained dual
coordinator from the coordination system this forecaster belongs
to~\citep{wang2026population}, which focuses the training distribution on cost
signals relevant to feasible capacity-control problems rather than arbitrary
perturbations in $\lambda$-space.

The simulator is then rolled out under each $\lambda^{(n)}_{0:T}$, applying the
broadcast costs to the fixed local policies and recording the induced aggregate
utilization $J^{(n)}_{0:T}$. This yields supervised training pairs
\[
(X_t^{(n)},\,\lambda^{(n)}_{t:t+L}) \;\mapsto\; J^{(n)}_{t:t+L},
\]
where at each training step $t$ the window $\lambda^{(n)}_{t:t+L}$ and the
corresponding target $J^{(n)}_{t:t+L}$ are extracted from the rollout.

\subsection{Training Procedure} \label{app:implementation}

At each training epoch, a capacity-target path $G_{0:T}\sim\mathcal{P}_G$ is
sampled from the wavelet distribution and mapped to a cost trajectory as above.
The response forecaster $\psi_\theta$ is trained by offline supervised regression on
the resulting rollouts, without backpropagating through the local policies:
\[
\hat{J}_{t:t+L}
=
\psi_\theta(X_t,\lambda_{t:t+L}),
\qquad
\mathcal{L}(\theta)
=
\sum_t
\left\|
\hat{J}_{t:t+L}-J_{t:t+L}
\right\|_2^2 .
\]
The local policies are frozen throughout; only $\psi_\theta$ is updated.

\subsection{Training Hyperparameters} \label{app:hyperparams}

Table~\ref{tab:hyperparams} lists the training hyperparameters for the response
forecaster.

\begin{table}[htbp]
\centering
\small
\caption{Training hyperparameters for the response forecaster $\psi_\theta$.}
\label{tab:hyperparams}
\begin{tabular}{lll}
\toprule
\textbf{Hyperparameter} & \textbf{Value} & \textbf{Notes} \\
\midrule
\multicolumn{3}{l}{\textit{Optimization}} \\
Optimizer & Adam & \\
Learning rate & $10^{-3}$ & Fixed throughout \\
Batch size & 3{,}000 & Products sampled per rollout; 5 cohorts per epoch\\
Cohort resampling & every 10 epochs & Uniform over the study population\\
Training epochs & 2{,}000 & \\
Gradient clipping & 0.01 ($\ell_\infty$-norm) &  \\
\midrule
\multicolumn{3}{l}{\textit{Sequence lengths}} \\
History window $H$ & 64 & Weeks of lookback \\
Forecast horizon $L$ & 26 & Weeks ahead \\
Training episode length   & 156  & Weeks per episode; start randomized \\
Evaluation horizon  &  52  &  Weeks per evaluation rollout; held out from training \\
\midrule
\multicolumn{3}{l}{\textit{Architecture}} \\
CNN channel width & 32 & Both history and future encoders \\
Future CNN channel width & 16 & \\
Decoder hidden size & 32 & Two-layer ELU MLP \\
History dilation rates & $[1,2,4,8,16,32]$ & Stacked causal CNN \\
Future dilation rates & $[1,2,4,8,11]$ & Reversed causal CNN \\
Encoder embedding $d_e$  &  32  &  Per-product encoder output \\
\bottomrule
\end{tabular}
\end{table}

\section{Additional Simulator Diagnostics} \label{app:fidelity_diagnostics}
The analyses in this appendix support the component-level fidelity results of Section~\ref{subsec:fidelity} with scatter diagnostics, an exogeneity check, a transition-logic validation, rollout-protocol sensitivity, and a policy-sensitivity analysis. The diagnostics are measured on Study~1, comparing simulated quantities against realized observations at the product--week level under the deployed policy $\pi$, using the calibration regression of Section~\ref{subsec:fidelity}. The regressions are fit by ordinary least squares without intercept over all product--week pairs of the Study~1 cohort in the evaluation window; reported intervals are the 95\% OLS intervals for the slope, and $R^2$ is the uncentered coefficient of determination of the no-intercept fit.

\subsection{Component-Level Transition Fidelity} \label{app:trajectory_fidelity}

\begin{figure}[htbp]
\centering
\includegraphics[width=\textwidth]{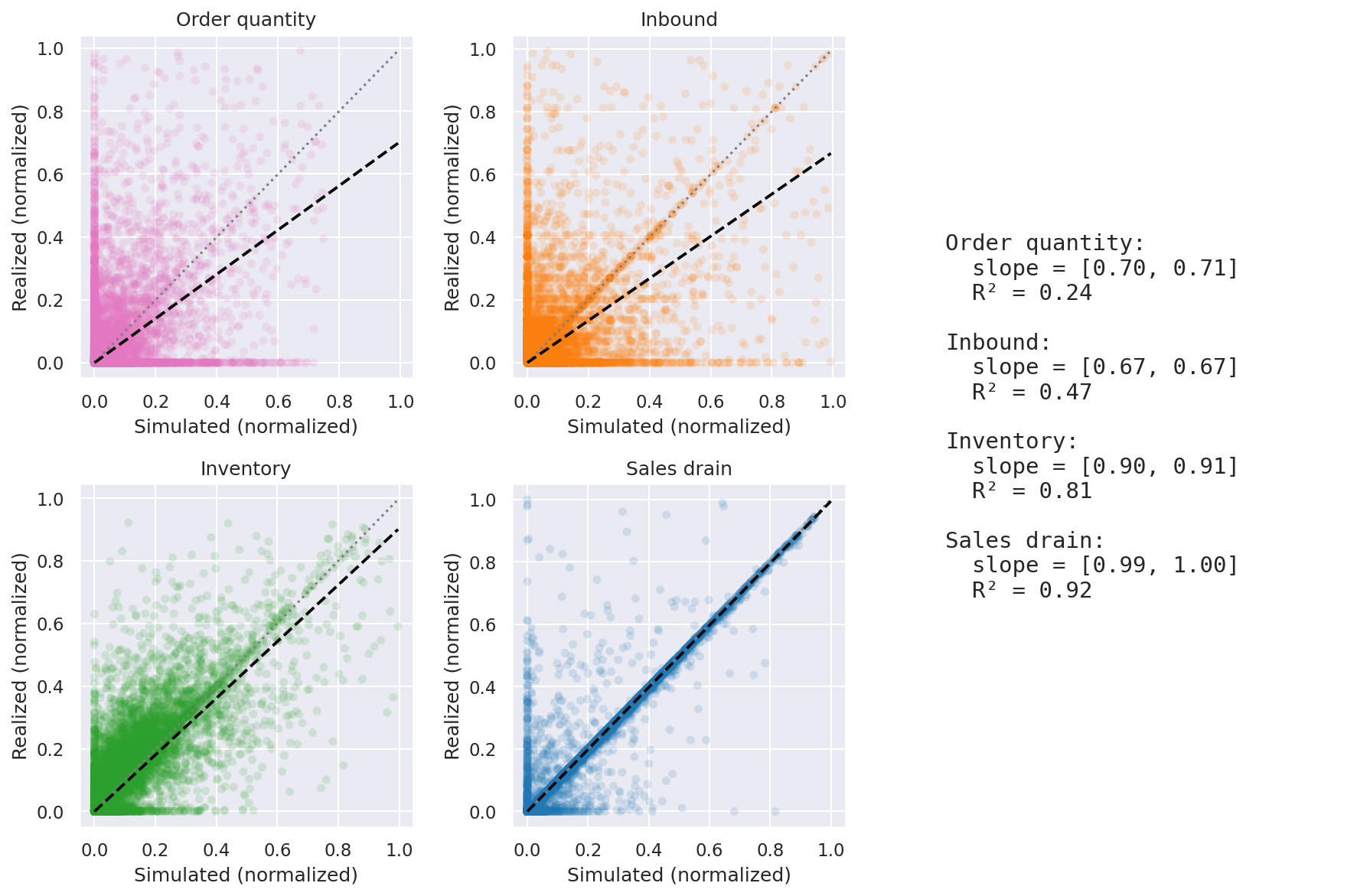}
\caption{\small Trajectory fidelity under the deployed policy: each point represents a product--week, with the $x$-axis showing simulated values and the $y$-axis showing corresponding realized observations; both axes of each panel are scaled to $[0,1]$ by a common constant, which leaves the regression slope and $R^2$ unchanged. A linear regression fit reports slope and $R^2$, with the dotted line indicating perfect calibration.}
\label{fig:trajectory_fidelity}
\end{figure}

Figure~\ref{fig:trajectory_fidelity} shows the product--week scatter underlying the calibration slopes reported in Table~\ref{tab:fidelity_components}, measured under the one-step rollout protocol (Appendix~\ref{app:rollout_protocol}). One plausible explanation for the order and inbound gaps is that production execution constraints modify orders but are not represented in the simulator; the diagnostic does not identify the cause. Because aggregate inbound is the forecasting target, product-level inbound calibration provides a finer-grained diagnostic of the simulator-generated target.

\subsection{Exogeneity Check} \label{app:exogeneity}

The replay argument of Section~\ref{subsec:task} assumes that exogenous processes evolve
independently of the ordering actions. Study~1's randomized design permits a direct test: for each
exogenous variable, we regress its observed value on the treatment indicator; under exogeneity, the
treatment coefficient should be indistinguishable from zero. Table~\ref{tab:exogeneity} reports the
estimates. Price and cost are consistent with exogeneity, while demand and vendor lead time show
statistically significant but small differences (both with very low $R^2$), indicating that the
assumption holds approximately but not perfectly and is worth further investigation.

\begin{table}[htbp]
\centering
\caption{\small Exogeneity check (Study~1): regression of representative exogenous variables on the
treatment indicator. Under exogeneity, coefficients should be indistinguishable from zero.}
\label{tab:exogeneity}
\footnotesize
\begin{tabular}{lcc}
\toprule
\textbf{Exogenous variable} & \textbf{Coef.\ (95\% CI)} & $R^2$ \\
\midrule
Demand           & $[1.37,\,2.44]$  & $0.02$ \\
Price            & $[-1.68,\,0.18]$ & $0.08$ \\
Cost             & $[-0.26,\,0.19]$ & $0.05$ \\
Vendor lead time & $[0.35,\,0.37]$  & $0.01$ \\
\bottomrule
\end{tabular}
\end{table}

\subsection{Transition-Logic Validation} \label{app:transition_validation}

To confirm that the calibration gaps in Table~\ref{tab:fidelity_components} arise from the learned
model components rather than implementation errors, we validate the simulator's deterministic
inventory-update logic: feeding realized orders, receipts, and sales into the inventory transition
equations of Section~\ref{subsec:task} reconstructs realized inventory with regression coefficient
$[0.95,\,0.96]$ and $R^2 = 0.948$. State propagation is therefore correctly implemented, and the
residual fidelity gaps are attributable to the learned order and inbound components.

\subsection{Rollout-Protocol Sensitivity} \label{app:rollout_protocol}

The diagnostics above use a one-step rollout strategy that reinitializes the simulator from realized
states at each evaluation week. Table~\ref{tab:rollout_protocol} compares this with a continuous
rollout that initializes once and simulates forward without resetting. Calibration slopes are broadly
similar across the two rollout protocols, and inventory and sales drain remain well calibrated under
both; pointwise agreement for inbound, however, decreases under continuous rollouts as stochastic
variation accumulates. Averaging over multiple independent rollouts reduces this variation for flow
quantities.

\begin{table}[htbp]
\centering
\caption{\small Rollout-protocol sensitivity (Study~1): product--week calibration under one-step
versus continuous rollouts. The one-step column corresponds to the diagnostics of
Table~\ref{tab:fidelity_components}.}
\label{tab:rollout_protocol}
\footnotesize
\begin{tabular}{lcc|cc}
\toprule
& \multicolumn{2}{c|}{\textbf{One-step}} & \multicolumn{2}{c}{\textbf{Continuous}} \\
\textbf{Component} & Coef.\ (95\% CI) & $R^2$ & Coef.\ (95\% CI) & $R^2$ \\
\midrule
Order quantity & $[0.70,\,0.71]$ & $0.24$ & $[0.65,\,0.66]$ & $0.25$ \\
Inbound        & $[0.67,\,0.67]$ & $0.47$ & $[0.65,\,0.66]$ & $0.12$ \\
Inventory      & $[0.90,\,0.91]$ & $0.81$ & $[1.04,\,1.04]$ & $0.87$ \\
Sales drain    & $[0.99,\,1.00]$ & $0.92$ & $[0.94,\,0.94]$ & $0.87$ \\
\bottomrule
\end{tabular}
\end{table}

\subsection{Policy Sensitivity} \label{app:policy_sensitivity}

\begin{figure}[htbp]
\centering
\includegraphics[width=0.9\textwidth]{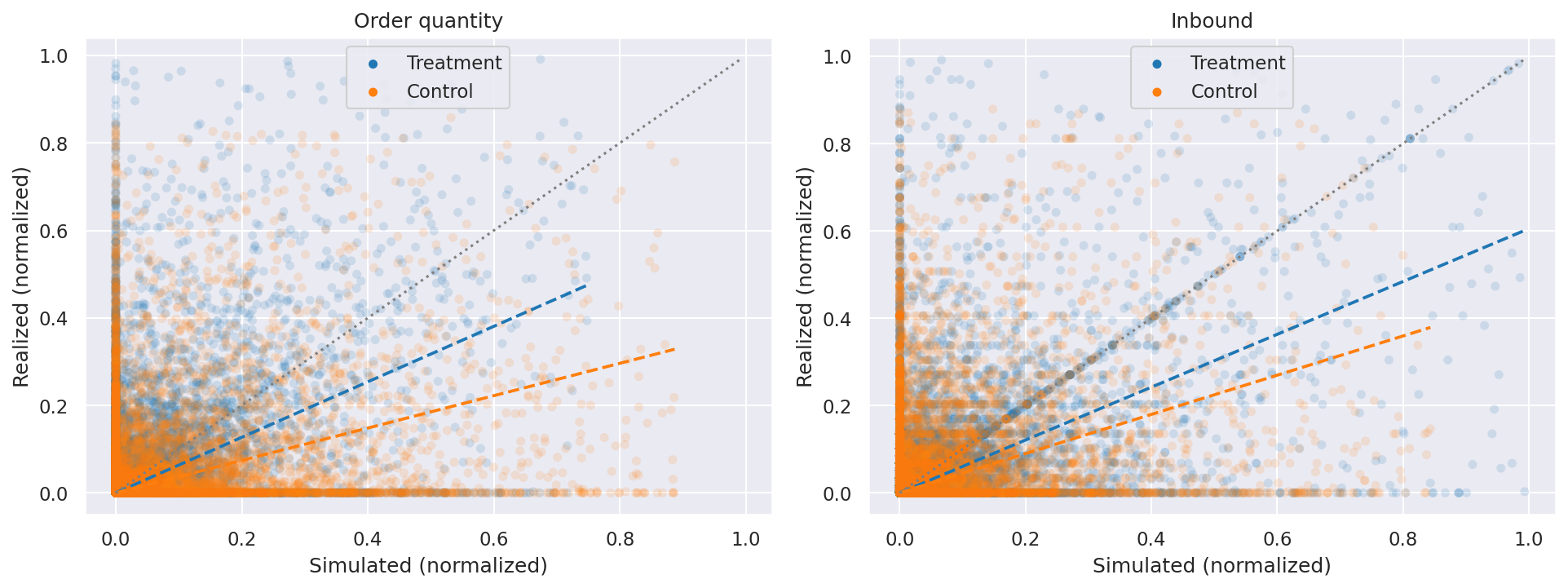}
\caption{\small Policy sensitivity: the $x$-axis shows simulated values under the deployed policy, the $y$-axis realized observations; both axes of each panel are scaled to $[0,1]$ by a common constant. Separate regressions for products operated under the deployed policy (on-policy) and under the baseline (off-policy); the dotted line indicates perfect calibration.}
\label{fig:trajectory_control}
\end{figure}

A simulator built from historical data must be assessed under the policy for which it will be used. We roll out the deployed policy for all products -- including those that operated under the baseline policy in reality -- and compare calibration on the two groups. For baseline products, the comparison is between the simulated deployed policy and realized baseline-policy outcomes, so part of the divergence is expected from the policy difference itself rather than simulator error. As shown in Figure~\ref{fig:trajectory_control}, calibration is stronger for products actually operated under the deployed policy (on-policy) than for those operated under the baseline (off-policy). This divergence is evidence that measured fidelity is policy-conditional. It does not by itself identify the mechanism causing the off-policy discrepancy, but it directly supports the premise of the main text: data collected under one policy need not characterize behavior under another, whether that data is real or simulated.

\section{Additional Transfer Results} \label{app:sim2real_detail}
This appendix provides full detail for the backtest results summarized in Section~\ref{sec:validation}: learnability of simulated dynamics, Sim2Real transfer, and calibration with early real observations.

\paragraph{Backtest data cutoffs.}
The simulator's transition and reward models and the forecaster $\hat{\psi}_{\mathrm{sim}}$ are fit before each deployment's evaluation window, whose realized outcomes are held out. Response-model training uses simulator rollouts from the pre-evaluation period. For the direct simulator-fidelity diagnostic, realized exogenous sequences from the evaluation window are replayed, so the measured error reflects the simulator's endogenous dynamics rather than exogenous forecast error. Adaptation follows the expanding-window protocol of Section~\ref{subsec:protocol}: calibration begins after the first four deployment weeks, and at each forecast origin uses only target-policy outcomes observed by that time, so no future labels enter the calibration. Zero-shot and calibrated variants are evaluated on the same forecast origins. Confidence intervals in the tables below are 95\% normal-approximation intervals across weekly forecast origins with one error observation per origin and horizon. Section~\ref{sec:related} discusses how this backtest design relates to hindcast evaluation and off-policy evaluation. The main text uses the Global Aggregate forecaster; as a robustness check, the tables below repeat available analyses with a Bottom-Up forecaster that sums per-product predictions.

\subsection{Learnability of Simulated Dynamics: Sanity Check} \label{app:sim2real_learnability}

Before the transfer evaluation, we check that simulator rollouts provide a stable supervised training signal at all. Models are trained on simulator rollouts and evaluated on temporally held-out simulator data from the same study; real-data references use the same model classes and training procedures. This is a Sim$\to$Sim diagnostic, not a transfer result. As shown in Table~\ref{tab:learnability}, simulator-trained models achieve stable errors on held-out simulator trajectories, with error ranges broadly comparable to those of models trained and evaluated on historical real data, for both model classes.

\begin{table}[htbp]
\centering
\caption{\small Learnability of simulator-generated dynamics. Models are trained on simulator rollouts and evaluated on held-out simulator data from the same study. Historical real-data baselines are trained on pre-study historical observations and evaluated on held-out real data from the Study~2 evaluation period, using the same model classes and training procedures. Reported values are MAPE (mean $\pm$ 95\% CI).}
\label{tab:learnability}
\resizebox{\textwidth}{!}{
\begin{tabular}{lllrrrr}
\toprule
\textbf{Training Source} & \textbf{Eval Source} & \textbf{Model} & \textbf{1 week} & \textbf{2 weeks} & \textbf{3 weeks} & \textbf{4--10 weeks} \\
\midrule
Simulator (Study~1) & Simulator (Study~1) & Global Aggregate & $9.2 \pm 1.8$ & $9.2 \pm 1.6$ & $10.8 \pm 2.0$ & $11.1 \pm 2.0$ \\
Simulator (Study~1) & Simulator (Study~1) & Bottom-Up & $12.2 \pm 3.1$ & $11.5 \pm 2.7$ & $10.7 \pm 2.5$ & $9.6 \pm 0.9$ \\
\midrule
Simulator (Study~2) & Simulator (Study~2) & Global Aggregate & $10.0 \pm 1.4$ & $10.5 \pm 1.5$ & $11.9 \pm 1.6$ & $9.1 \pm 1.3$ \\
Simulator (Study~2) & Simulator (Study~2) & Bottom-Up & $10.0 \pm 2.6$ & $12.9 \pm 2.8$ & $14.0 \pm 3.0$ & $12.2 \pm 1.7$ \\
\midrule
Historical Real Data & Held-out Real Data & Global Aggregate & $11.7 \pm 1.5$ & $11.9 \pm 1.5$ & $11.8 \pm 1.4$ & $11.5 \pm 1.3$ \\
Historical Real Data & Held-out Real Data & Bottom-Up & $8.9 \pm 2.2$ & $9.9 \pm 2.3$ & $10.5 \pm 2.4$ & $9.7 \pm 1.2$ \\
\bottomrule
\end{tabular}}
\end{table}

\subsection{Zero-Shot Transfer: Full Results} \label{app:sim2real_transfer}

Table~\ref{tab:sim2real_transfer} reports the available horizon-resolved zero-shot transfer results summarized in Section~\ref{subsec:transfer}. The source results do not include an off-policy Bottom-Up row for Study~2, so that comparison is not reported.

\paragraph{Benchmark construction.}
In Study~1 the historical benchmark shares the simulator-trained forecaster's architecture, is fit to realized observations of the same product cohort from the weeks preceding the deployment, and is scored on the same series at the same forecast origins. Study~2 follows a similar construction.

\begin{table}[htbp]
\centering
\caption{\small Zero-shot transfer, full results. Models are trained on pre-evaluation simulator rollouts under the deployed policy or on historical real data collected under prior operating policies (off-policy), and evaluated on realized aggregate utilization during the matched deployment windows. Reported values are MAPE (mean $\pm$ 95\% CI) by forecast horizon.}
\label{tab:sim2real_transfer}
\resizebox{\textwidth}{!}{
\begin{tabular}{lllrrrr}
\toprule
\textbf{Training Source} & \textbf{Eval Source} & \textbf{Model} & \textbf{1 week} & \textbf{2 weeks} & \textbf{3 weeks} & \textbf{4--10 weeks} \\
\midrule
Simulator (Study~1) & Real (Study~1) & Global Aggregate & $13.3\pm 2.7$ & $11.7\pm 2.8$ & $13.2\pm 3.1$ & $14.8\pm 2.8$ \\
Simulator (Study~1) & Real (Study~1) & Bottom-Up & $13.4\pm 3.0$ & $13.3\pm 3.0$ & $13.4\pm 2.8$ & $13.4\pm 1.1$ \\
\midrule
Simulator (Study~2) & Real (Study~2) & Global Aggregate & $12.7\pm 3.6$ & $15.3\pm 7.4$ & $9.4\pm 3.8$ & $11.5\pm 3.4$ \\
Simulator (Study~2) & Real (Study~2) & Bottom-Up & $16.4\pm 6.2$ & $17.7\pm 7.4$ & $20.1\pm 8.0$ & $19.4\pm 3.9$ \\
\midrule
Off-policy Real Data & Real (Study~1) & Global Aggregate & $15.0\pm 2.9$ & $14.8\pm 3.1$ & $15.2\pm 3.1$ & $16.0\pm 3.1$ \\
Off-policy Real Data & Real (Study~1) & Bottom-Up & $16.7\pm 2.9$ & $16.8\pm 3.0$ & $17.7\pm 3.0$ & $19.5\pm 1.2$ \\
Off-policy Real Data & Real (Study~2) & Global Aggregate & $25.6 \pm 5.2$ & $27.8 \pm 6.3$ & $28.1 \pm 6.4$ & $25.2 \pm 5.7$ \\
\bottomrule
\end{tabular}}
\end{table}

\subsection{Sim2Real Transfer with Real-Data Adaptation} \label{app:sim2real_adaptation}

We apply lightweight calibration using observations collected early in the real-world evaluation period. For each model, we fit a simple regression adjustment from model forecasts to realized inbound outcomes, and then evaluate the calibrated forecasts on the remaining evaluation window.

Table~\ref{tab:sim2real_adaptation} reports the available forecasting results after calibration. The source results do not include a calibrated off-policy Global Aggregate row for Study~2; conversely, its calibrated off-policy Bottom-Up row has no retained zero-shot counterpart. We therefore draw before--after conclusions only where both variants are available. Calibration lowers point-estimate error for the simulator-trained Global Aggregate model in both studies, most at horizons of three weeks or less, and for the Bottom-Up model at longer horizons, but effects are not uniform across models or studies. These results characterize the observed correction behavior but do not identify its cause.

\begin{table}[htbp]
\centering
\caption{\small Available forecasting results after regression-based calibration using early real observations. Simulator-trained models are corrected using data from the beginning of the evaluation period. Off-policy historical baselines use the same calibration procedure. The retained source results do not contain every model--study combination. Values are MAPE (mean $\pm$ 95\% CI).}
\label{tab:sim2real_adaptation}
\resizebox{\textwidth}{!}{
\begin{tabular}{lllrrrr}
\toprule
\textbf{Training Source} & \textbf{Eval Source} & \textbf{Model} & \textbf{1 week} & \textbf{2 weeks} & \textbf{3 weeks} & \textbf{4--10 weeks} \\
\midrule
Simulator (Study~1) & Real (Study~1) & Global Aggregate + Calib. & $11.8\pm 2.6$ & $11.2\pm 3.1$ & $13.1\pm 3.6$ & $14.8\pm 3.2$ \\
Simulator (Study~1) & Real (Study~1) & Bottom-Up + Calib. & $13.5 \pm 6.0$ & $12.4 \pm 5.7$ & $12.3 \pm 5.4$ & $11.2 \pm 1.8$ \\
\midrule
Simulator (Study~2) & Real (Study~2) & Global Aggregate + Calib. & $12.6 \pm 3.6$ & $13.0 \pm 7.1$ & $6.9 \pm 3.1$ & $9.7 \pm 2.9$ \\
Simulator (Study~2) & Real (Study~2) & Bottom-Up + Calib. & $19.6 \pm 6.8$ & $21.3 \pm 6.3$ & $18.2 \pm 9.1$ & $14.4 \pm 4.7$ \\
\midrule
Off-policy Real Data & Real (Study~1) & Global Aggregate + Calib. & $15.5\pm 3.0$ & $15.9\pm 3.3$ & $15.8\pm 3.6$ & $18.0\pm 3.4$ \\
Off-policy Real Data & Real (Study~1) & Bottom-Up + Calib. & $18.4\pm 3.0$ & $15.4\pm 2.6$ & $15.2\pm 2.3$ & $14.1\pm 1.8$ \\
Off-policy Real Data & Real (Study~2) & Bottom-Up + Calib. & $29.9 \pm 9.6$ & $29.6 \pm 11.7$ & $28.4 \pm 12.9$ & $20.5 \pm 5.5$ \\
\bottomrule
\end{tabular}}
\end{table}

\paragraph{Summary.}
Taken together, these results show that the simulator has measurable aggregate discrepancy yet generates learnable trajectories and response forecasters whose point-estimate errors on realized outcomes are lower than the corresponding off-policy historical baselines in most reported comparisons. Lightweight calibration reduces residual error in some settings but not uniformly. These conclusions are empirical, conditional on the two deployments and the simulator used here, and do not identify data source or policy alignment as the causal mechanism.

\end{document}